\documentclass[preprint,12pt]{elsarticle}
\usepackage{amssymb}
\usepackage{lipsum}
\usepackage{hyperref}
\usepackage{amsmath}
\usepackage{booktabs}
\usepackage{amsmath}
\usepackage{graphicx} 
\usepackage{xcolor}

\usepackage{xcolor} 
\usepackage{hyperref} 

\journal{High Energy Astrophysics}

\begin{document}

\begin{frontmatter}

\title{EA-3DGS: Efficient and Adaptive 3D Gaussians with Highly Enhanced Quality for outdoor scenes}


\author[SCUT,FN1]{Jianlin Guo}
\ead{202321016915@mail.scut.edu.cn}
\author[SCUT,FN1]{Haihong Xiao}
\ead{auhhxiao@mail.scut.edu.cn}
\author[SCUT]{Wenxiong Kang\corref{COR1}}
\ead{auwxkang@scut.edu.cn}

\affiliation[SCUT]{organization={South China University of Technology}, 
            city={China},
            country={Guangzhou}}

\begin{abstract}
Efficient scene representations are essential for many real-world applications, especially those involving spatial measurement. Although current NeRF-based methods have achieved impressive results in reconstructing building-scale scenes, they still suffer from slow training and inference speeds due to time-consuming stochastic sampling.
Recently, 3D Gaussian Splatting (3DGS) has demonstrated excellent performance with its high-quality rendering and real-time speed, especially for objects and small-scale scenes. However, in outdoor scenes, its point-based explicit representation lacks an effective adjustment mechanism, and the millions of Gaussian points required often lead to memory constraints during training. To address these challenges, we propose EA-3DGS, a high-quality real-time rendering method designed for outdoor scenes. First, we introduce a mesh structure to regulate the initialization of Gaussian components by leveraging an adaptive tetrahedral mesh that partitions the grid and initializes Gaussian components on each face, effectively capturing geometric structures in low-texture regions. Second, we propose an efficient Gaussian pruning strategy that evaluates each 3D Gaussian's contribution to the view and prunes accordingly. To retain geometry-critical Gaussian points, we also present a structure-aware densification strategy that densifies Gaussian points in low-curvature regions. Additionally, we employ vector quantization for parameter quantization of Gaussian components, significantly reducing disk space requirements with only a minimal impact on rendering quality. Extensive experiments on 13 scenes, including eight from four public datasets (MatrixCity-Aerial, Mill-19, Tanks \& Temples, WHU) and five self-collected scenes acquired through UAV photogrammetry measurement from SCUT-CA and plateau regions, further demonstrate the superiority  of our method. Codes are available at \href{https://github.com/SCUT-BIP-Lab/EA-3DGS}{https://github.com/SCUT-BIP-Lab/EA-3DGS}.
\end{abstract}


\begin{graphicalabstract}
\centering
\includegraphics[width=.9\textwidth]{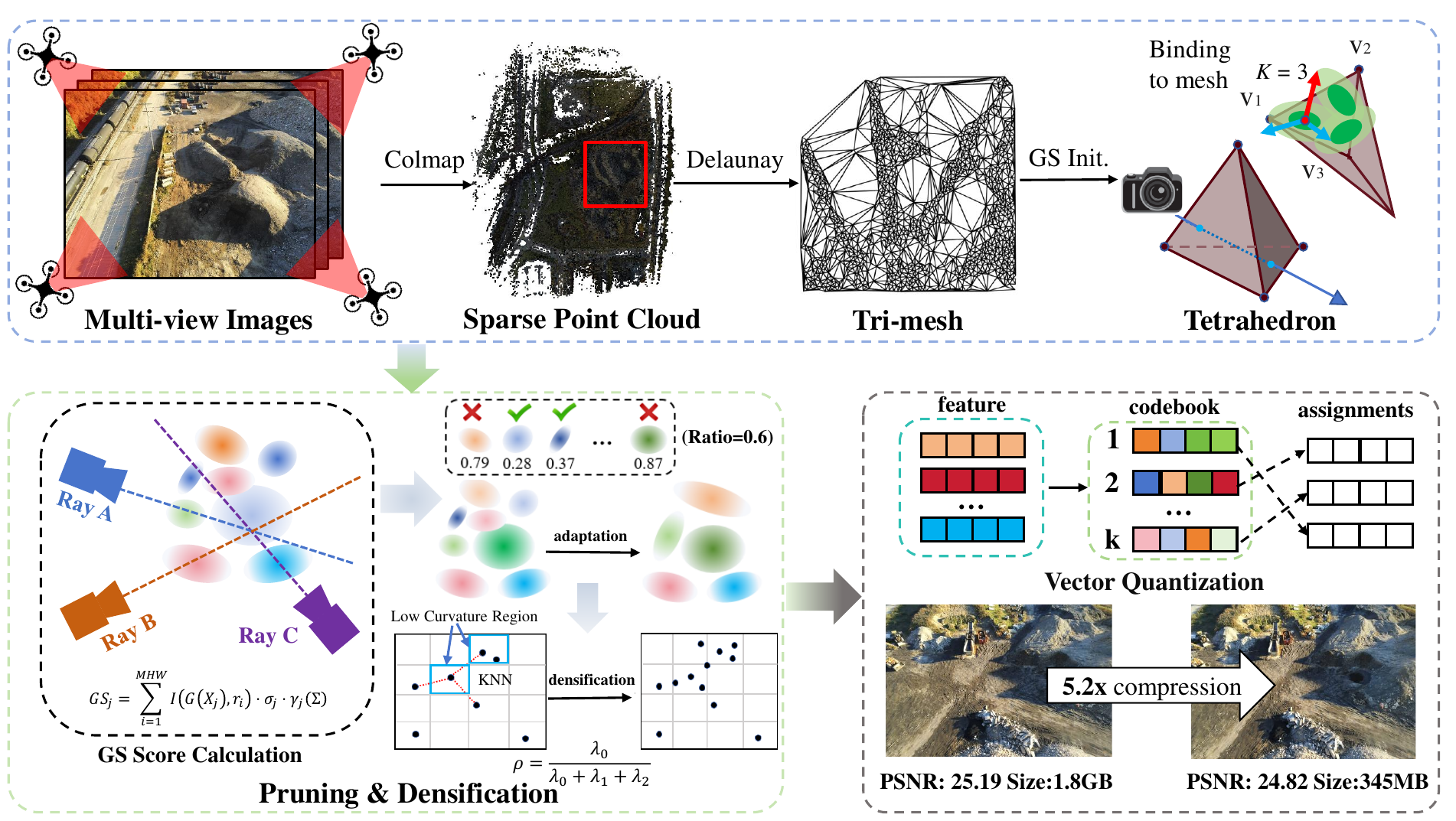}
\end{graphicalabstract}

\begin{highlights}
\item Adaptive mesh guides 3D Gaussian initialization for outdoor scene geometry.
\item Contribution-aware pruning \& structure-aware densification optimize 3D Gaussians.
\item Codebook quantization enables compact, high-quality 3DGS for real-time rendering.
\end{highlights}


\cortext[COR1]{Corresponding author.}
\fntext[FN1]{These authors contributed equally to this work.}

\end{frontmatter}




\section{Introduction}
\label{introduction}
\begin{figure*}
\centering
\includegraphics[width=.9\textwidth]{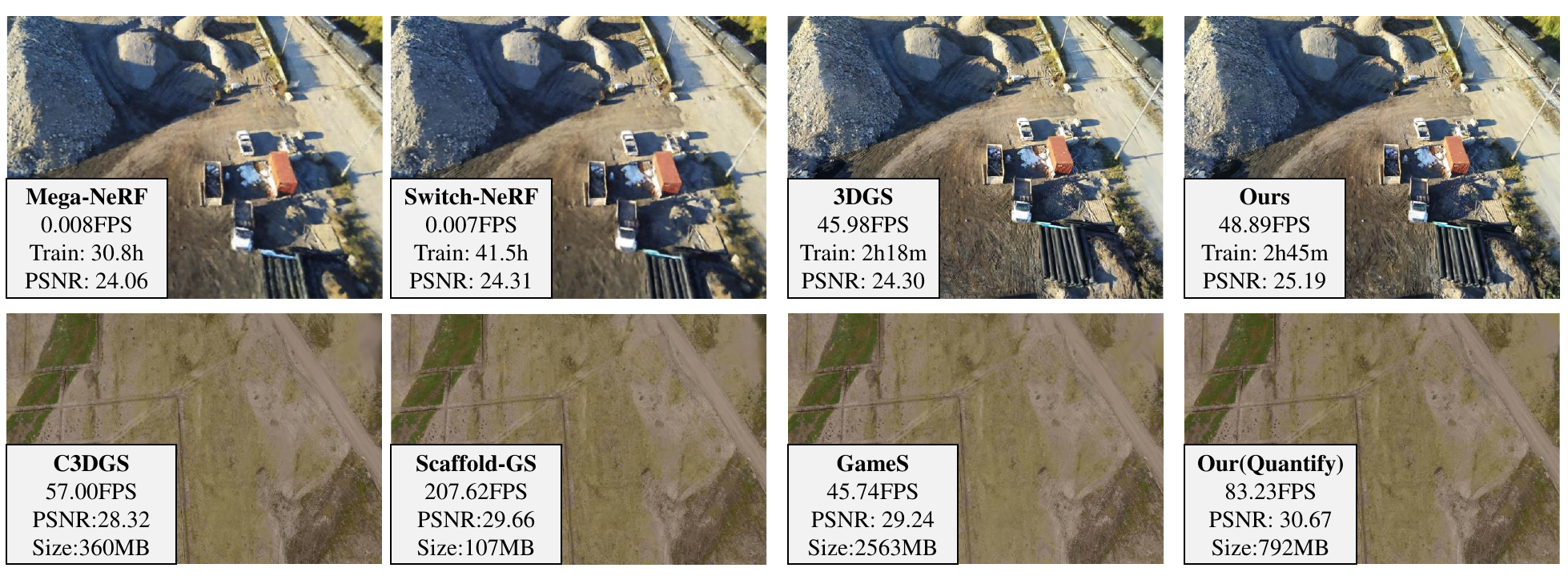}
\caption{Compared to previous NeRF-based methods (such as Mega-NeRF and Switch-NeRF) and 3DGS-based methods (including 3DGS, C3DGS, Scaffold-GS, and GaMeS), we observe that Mega-NeRF \cite{turki2022mega} and Switch-NeRF \cite{zhenxing2022switch} display noticeable blurriness and slow rendering speeds. All comparisons were conducted on an NVIDIA 3090 GPU with 24GB of memory. Relevant parameters are shown in the figure.}
\label{fig:front}
\end{figure*}
Equipping machines with the ability to reconstruct and understand three-dimensional scenes has long been a tantalizing pursuit, promising a bridge between the digital and physical worlds. 3D reconstruction has attracted considerable interest from academia and industry alike, driven by the need for high-fidelity, real-time rendering of large-scale scenes such as cities and terrains. This technology provides essential support for applications including autonomous driving \cite{gomez2023three}, geological exploration \cite{zhang2025learning}, and virtual reality \cite{zhang2025c2fi}.

In the early stages, 3D reconstruction was predominantly based on sparse Structure-from-Motion (SfM) \cite{Relate_sfm,Relate_colmap} and dense Multi-View Stereo (MVS) \cite{gu2024ea,xu2023semi} pipelines. However, recent advances have been dominated by Neural Radiance Fields (NeRF) \cite{nerf}, an emerging technique for generating metrically consistent 3D models from posed 2D images. Although NeRF and its variants \cite{hu2024td,xie2025tri} have recently demonstrated impressive performance in reconstructing objects or small scenes from images, directly extending these methods to larger outdoor scenes is not straightforward. Intuitively, larger scenes require larger networks, which could be achieved by increasing the network’s width and depth to accommodate the expanded scene representation. However, such a naive operation inevitably incurs substantial computational costs and poses significant optimization challenges. Representative works include Block-NeRF\cite{tancik2022block}, Switch-NeRF\cite{zhenxing2022switch}, and BungeeNeRF\cite{xiangli2022bungeenerf}, which introduce partitioning, parallel computation strategies and select different network parameters (\textit{i.e.}, sub-networks) to improve feasibility in outdoor scenes. However, they inherently inherit the drawbacks of NeRF and still struggle with limitations in detailed fidelity and rendering speed.

3D Gaussian Splatting (3DGS) \cite{3dgs} has opened up a new research direction for scene reconstruction. In contrast to NeRF's implicit representation, 3DGS employs explicit 3D Gaussian functions as primitives for scene rendering. The basic idea is to represent a scene as a large unstructured collection of fuzzy particles which can be differentiably rendered by splatting to a camera view with a tile-based rasterizer. It has rapidly sparked a plethora of studies \cite{10926911, turkulainen2024dn}, ranging from objects to small-scale scenes, and from static to dynamic scenes. 

However, challenges persist when training on large-scale scenes. Directly applying 3DGS to such scenes leads to excessive GPU memory consumption during training. The explicit representation triggers out-of-memory errors on a $24\ \text{GB}$ RTX 3090 GPU when the number of Gaussians reaches millions, interrupting the training process. Furthermore, numerous studies \cite{li2024ho} indicate that the performance of 3DGS heavily relies on well-initialized point clouds from Structure-from-Motion (SfM)\cite{sfm} or Simultaneous Localization and Mapping (SLAM)\cite{wang2025eum, wu2024kn}. Although random initialization performs adequately for small scenes, it proves insufficient for large-scale scenes with limited fields of view and unbounded regions, often misinterpreting uniformly textured areas as outliers and resulting in incomplete 3D reconstructions.

In this work, we validate our method on large-scale scenes, addressing two key challenges previously underexplored:
\begin{enumerate}
\item{3D Gaussian Splatting (3DGS) suffers from a deficiency in geometric guidance, leading to models with inadequate geometric consistency and limited scene structure understanding.}
\item{Training unbounded, large-scale real-world scenes incurs prohibitive costs in computation and storage, lacking a generalized and efficient solution for scalable training.}
\end{enumerate}

To address the aforementioned challenges and enable real-time rendering of 3DGS in large-scale scenes, we propose an efficient and adaptive 3DGS approach, EA-3DGS. Due to the presence of weak textures, repetitive textures, and low overlaps between views, the initial point clouds often contain missing points, which directly impact rendering quality. Inspired by methods such as Adaptive Shell\cite{adaptiveshell}, Binary Occupancy Field\cite{binaryfiled}, Scaffold-GS\cite{lu2024scaffold}, Sugar\cite{sugar}, and HAC \cite{chen2024hac}, we introduce geometry guidance to significantly enhance rendering quality. To this end, we innovatively propose an adaptive tetrahedral mesh to guide 3DGS, binding Gaussian functions to mesh surfaces and controlling their initialization through mesh constraints, enabling the model to better capture geometric information in low-texture areas. Additionally, we adopt a pruning strategy similar to LightGaussian, pruning irrelevant Gaussian points by calculating importance scores, which reduces the number of Gaussians during training and facilitates efficient training on standard GPU devices. To retain essential geometry-related Gaussian points, we further introduce a curvature-based densification strategy. Finally, we leverage the codebook\cite{codebook} technique to encode Gaussian components, significantly reducing the disk space required for Gaussian splatting. After compression, the model’s disk storage is reduced to one-fifth of its original size at the sacrifice of a minimal rendering visual cost, greatly improving storage efficiency and enabling the rendering of unbounded outdoor scenes. We show the comparison results in some scenarios in Fig. \ref{fig:front}.

\begin{figure}
    \centering
    \includegraphics[width=1\linewidth]{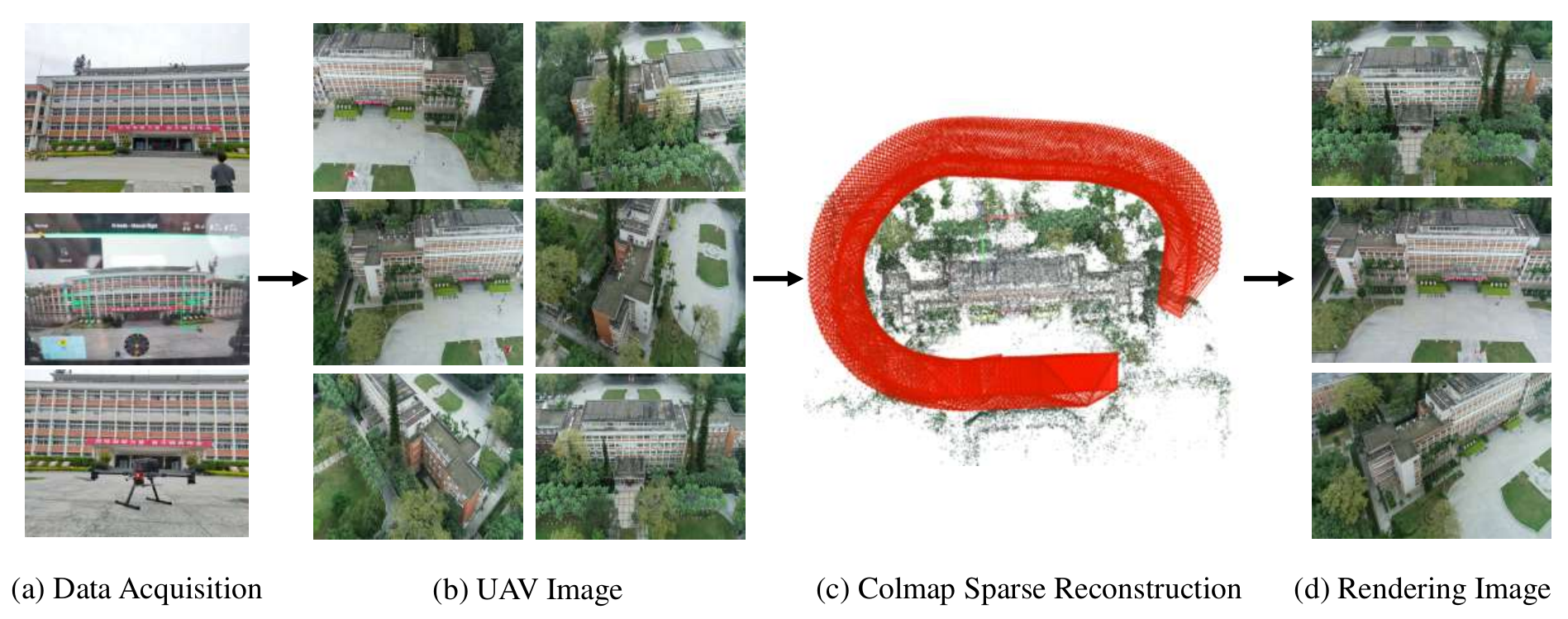}
    \caption{Data acquisition at the SCUT campus using a DJI Matrice 300 RTK UAV.}
    \label{fig:dataset}
\end{figure}

To summarize, our contributions are four-fold as follows.
\begin{itemize}
  \item We propose an adaptive tetrahedral mesh-guided initialization for 3DGS, binding Gaussians to surfaces via geometric constraints to effectively capture geometric information in low-texture areas.
  \item We design a dual Gaussian optimization: contribution-aware pruning reduces memory footprint, while structure-aware densification preserves geometry in low-curvature regions.
  \item We implement codebook-based parameter quantization for Gaussians, achieving disk compression with minimal rendering quality degradation.
  \item We collected five diverse UAV photogrammetry scenes (urban/natural terrain) from SCUT campus and plateau regions, the data collection process for which is illustrated in Fig. \ref{fig:dataset}. Experiments on 13 scenes demonstrate our method's effectiveness compared to state-of-the-art approaches.
\end{itemize}

\section{Related Work}

\subsection{Neural Radiance Field}
Neural Radiance Fields (NeRF)\cite{nerf} introduce a paradigm for representing a scene as an emissive volume, employing a neural network with position encoding that is rendered differentiably using quadrature and optimized through gradient descent. 
Despite their remarkable rendering capabilities\cite{kulhanek2023tetra, nopenerf}, NeRF is hampered by the substantial cost and time required to gather dense image sets for a given scene. This challenge is further amplified in outdoor scenes, a highly challenging domain focused on novel view synthesis of outdoor scenes. Mega-NeRF \cite{turki2022mega} partitions training images into different NeRF submodules to enable parallel training, leading to linear scale-up in training costs and the number of sub-NeRFs as the scene expands. Block-NeRF \cite{tancik2022block} adopts a block-based approach, decomposing large scenes into independently trained blocks and leveraging their spatial distribution for view synthesis. In contrast to these partitioning methods, GP-NeRF \cite{li2024gp} employs a hybrid feature representation that combines 3D hash grids with high-resolution 2D dense plane features. However, the feature grid tends to be less constrained and often produces noisy artifacts in renderings, especially in regions with complex geometry and texture. SceneRF \cite{cao2023scenerf} optimizes the radiance field through explicit depth optimization and a probabilistic sampling strategy, thereby enabling self-supervised monocular depth estimation for the reconstruction of complex scenes. Grid-NeRF \cite{xu2023grid} integrates the MLP-based NeRF with an compact multi-resolution grid to encode both local and global scene informations. Although this joint learning method enhances scene rendering quality, it still inherits some limitations of NeRF-based methods, such as the slow training speed. In addition, the batch sampling of shuffled rays is highly ineffective without distributed training. 

NeRF-based methods suffer from well-known problems, such as time-consuming rendering process, even though the most advanced NeRF techniques currently available cannot achieve real-time rendering in high-resolution, unrestricted scenes. 

\subsection{3D Gaussian Splatting}

3D Gaussian Splatting (3DGS) \cite{3dgs} addresses the challenges of slow training and rendering in NeRF by employing an explicit point-based reconstruction. This approach represents 3D scenes with 3D Gaussian functions featuring learnable shape and appearance attributes, allowing for differentiable rendering by splatting to a camera view with a tile-based rasterizer. Currently, 3DGS has been widely explored, ranging from object-level to scene-level reconstruction, and from static to dynamic scenes.

To achieve high-fidelity reconstruction and real-time rendering of large-scale scenes, most methods have adopted the divide-and-conquer strategy \cite{stearns2024dynamic, zhu2021paircon}. VastGaussian \cite{lin2024vastgaussian} proposes a progressive partitioning method  that assigns training views and point clouds to separate units, enabling parallel optimization and seamless fusion. DoGaussian \cite{chen2024dogaussian} introduces a distributed training method by partitioning the scene into multiple blocks and incorporating the Alternating Direction Method of Multipliers (ADMM) for optimization. The global information is maintained at the master node, while local scene information is managed at subordinate nodes. During inference, only the global 3DGS model is queried. CityGaussian \cite{liu2024citygaussian} 
employs a novel divideand-conquer training strategy and Level-of-Detail (LoD) method to achieve fast rendering across different scales. These methods primarily follow a partitioning approach, but managing multiple partitions and ensuring their correct synchronization and merging is very complex. This not only requires additional setup and configuration steps but also increases system complexity and potential points of failure. Moreover, ensuring training consistency across different partitions is challenging, as global information, such as lighting, may be lost during the merging process. In contrast, single-GPU training does not involve these cumbersome steps, allowing for a more direct handling of the entire scene and avoiding many of the issues associated with partitioning.

Moreover, some studies have considered incorporating structured information into high-quality rendering. GSDF \cite{yu2024gsdf} employs a dual-branch network that integrates Signed Distance Fields (SDF) to guide 3DGS, achieving density control and more detailed surface reconstruction. GaMeS \cite{waczynska2024games} utilizes a grid structure to guide 3DGS, parameterizing each Gaussian component with the vertices of the grid faces, resulting in an editable and efficient Gaussian model. While GSDF and GaMeS have made significant strides in incorporating structured information for high-quality rendering, they still face certain limitations. The dual-branch network of GSDF, although effective for density control and detailed surface reconstruction, can be computationally intensive and may struggle with complex geometries. In contrast, GaMeS relies on a grid structure that, while efficient and editable, may not fully capture the intricacies of large-scale scenes, particularly in areas with high variability.

Different from previous work, this paper adopts an adaptive tetrahedral mesh that parameterizes Gaussian distributions on each mesh face to guide the initialization of the Gaussian pipeline and enhance rendering results in large-scale scenes. Additionally, we propose an effective pruning and densification strategy to ensure stable training on single-GPU in large-scale scenarios.

\subsection{Compression for 3D scene representation methods}
The explicit storage of 3D Gaussian attributes and gradients has a notable drawback—it requires substantial memory and storage, especially in large-scale reconstructions. Rich details in large scenes typically necessitates numerous 3D Gaussian functions and directly applying 3D Gaussian splatting to such scenes can lead to memory insufficiency. Consequently, efficient compression techniques, such as pruning \cite{lp-3dgs}, codebooks \cite{compact-3dgs,xiao2024instance} and entropy constraints \cite{wang2024end}, are necessary. LightGaussian \cite{fan2023lightgaussian} employs pruning, SH distillation, and VTree quantization during training, while ScaffoldGS \cite{lu2024scaffold} utilizes anchor points to distribute local 3D Gaussians, reducing the parameter count and dynamically predicting their attributes based on the viewing frustum and observation distance. We observe that adjacent Gaussians often exhibit high similarity in their parameters. To address this, we introduce a novel approach that leverages Vector Quantization (VQ) technique to transform a large set of vectors into a smaller codebook, which serves as a collection of representative vectors used to quantize each original vector. This compression technique has been successfully applied across various domains, including image compression, audio-visual coding, feature extraction, and generative models. Unlike previous work, we apply the VQ technique to compress Gaussian parameters, thereby reducing model storage requirements while maintaining the quality and rendering speed of 3D Gaussian splatting.

\begin{figure*}[!t]
\centering
\includegraphics[width=5.5 in]{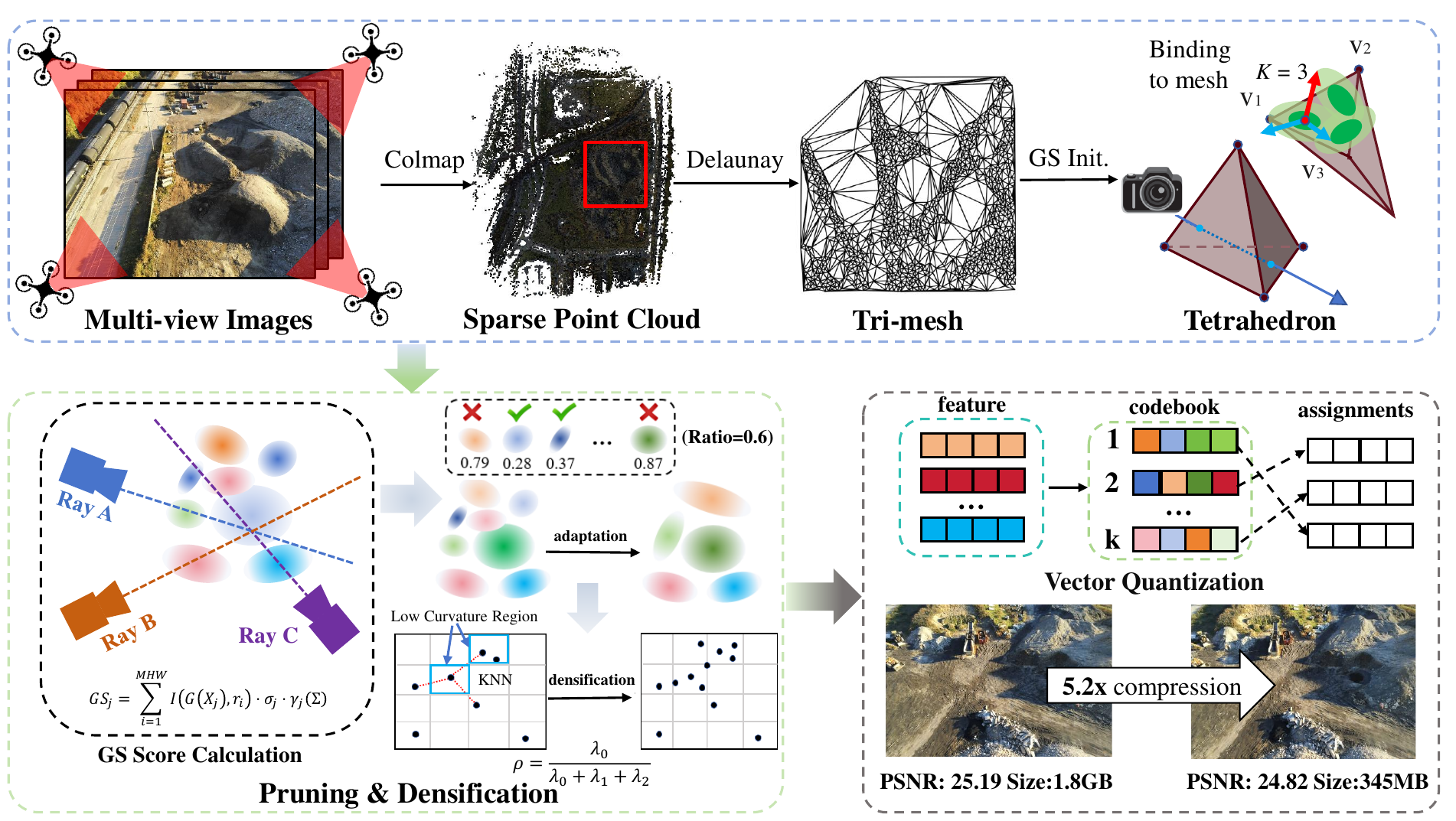}
\caption{\textbf{The overall pipeline of EA-3DGS.} Our proposed method begins by generating an adaptive tetrahedral mesh from the input multi-view images, which serves as the foundation for optimizing 3D Gaussians. During training, EA-3DGS assesses the global importance of each Gaussian distribution based on training observations, enabling the pruning of less significant Gaussians. To mitigate issues arising from excessive pruning, we introduce a structure-aware densification strategy that selectively increases Gaussian density in low-curvature regions. Furthermore, to facilitate real-time rendering in larger scenes, we apply the codebook technique to quantize Gaussian component parameters, effectively reducing memory usage.}
\label{fig:main}
\end{figure*}

\section{Method}

In this paper, we present an enhanced version of the 3DGS model for large-scale scene representation. The original 3DGS model is trained on multi-view posed images and initialized from Structure-from-Motion (SfM) point clouds, representing the scene by expanding a sparse point cloud into millions of Gaussian points. In contrast to this SfM-based initialization \cite{sfm}, our proposed EA-3DGS approach utilizes an adaptive tetrahedral representation to partition the grid, initializing Gaussian components on each grid face. This adaptive representation enables better capture of fine-grained scene details. Furthermore, to reduce GPU memory (VRAM) consumption while better preserving the scene’s geometric structure, we introduce a Gaussian pruning and structure-aware densification strategy in the training process. Finally, we further integrate the vector quantization technique into our approach, enhancing its competitiveness and suitability for outdoor scenes with minimal compromise. Experimental results demonstrate that our improved 3DGS model, EA-3DGS, exhibits superior results for large-scale scene modeling tasks while also achieving a smaller storage footprint compared to the original 3DGS.The pipeline is shown in Fig. \ref{fig:main}.

\subsection{3D Gaussian Splatting}
3DGS \cite{3dgs} is an explicit point-based representation for 3D scenes that utilizes various Gaussian attributes to model the scene. To address the complexities of real-world scene modeling, 3DGS initializes Gaussian components from the sparse point cloud generated by Structure-from-Motion (SFM) and adopts an adaptive density control mechanism to split or clone the points during the optimization process. Formally, the attributes of a 3D Gaussian include its center (position) $\mu$, opacity $\alpha$, 3D covariance matrix $\Sigma$, and color $c$, with the color $c$ represented by spherical harmonics (SH) to achieve view-dependent appearance.
\begin{equation}
    G(\chi) = e^{-\frac{1}{2}(\chi)^{T} \Sigma^{-1} (\chi) },
\end{equation}
where $\chi$ is the position centered at the mean point $\mu$, and $\Sigma$ is the 3D covariance matrix of the Gaussian. Since the covariance matrix needs to be positive semi-definite, $\Sigma$ can be decomposed into a scaling matrix $S$ and a rotation matrix $R$ as $\Sigma = RSS^{T}R^{T}$ for differentiable optimization.

In the rendering process, we project 3D Gaussian ellipsoids onto 2D images. This involves a splatting technique, where the 3D Gaussians (ellipsoids) are projected into the 2D image space (ellipses) for rendering. Given the view transformation $W$ and the 3D covariance matrix $\Sigma$, the projected 2D covariance matrix ${\Sigma}'$ can be calculated using the following formula.
\begin{equation}
        {\Sigma}' = JW\Sigma W^{T}J^{T},
\end{equation}
where $J$ is the Jacobian of the affine approximation of the projective transformation.


Each Gaussian component has an associated opacity $\alpha$. From a given viewpoint, for a specific pixel, we can calculate the distances to all overlapping Gaussians, forming a depth-sorted list of Gaussians $N$. The color of that pixel can then be obtained using the following formula.
\begin{equation}
  C = \sum_{i\in N}^{} c_i {\alpha _i}' \prod_{j = 1}^{i - 1}(1-{\alpha _j}'),
\end{equation}
where $c_i$ represents the learned color of the Gaussian component, and ${\alpha_i}'$ is computed as the product of the learned opacity $\alpha_i$ and a function of the 3D covariance matrix $\Sigma$ and the spherical harmonics (SH).
\subsection{Adaptive Tetrahedra Mesh For Initialization}
The original 3D Gaussian splatting method adaptively controls the number of Gaussian points through cloning and splitting operations. This allows for the conversion of the initial sparse set into a denser set. However, this mechanism relies heavily on the initial point cloud’s quality and cannot effectively grow points in areas where the initial point cloud is sparse, resulting in blurry or needle-like artifacts in the synthesized images\cite{xiao2023distinguishing}. To address this limitation, we implement a parameterized Gaussian distribution on each face of an adaptive tetrahedral mesh, which provides new positions for the Gaussian representation.

Given a set of points in three-dimensional space, our goal is to construct a tetrahedral structure by triangulating these points. Consider a planar point set $P = {P_1, P_2, \ldots, P_n}$; we aim to obtain a set of tetrahedra $T = {t_1, t_2, \ldots, t_m}$ that satisfies the following conditions: a) The vertices of all the tetrahedra exactly constitute the set $P$. b) Any two tetrahedra do not intersect. c) The union of all the tetrahedra forms the convex hull of the point set $P$.

Inspired by \cite{kulhanek2023tetra,Yu2024GOF}, We employ Delaunay triangulation to generate a set of non-overlapping tetrahedra whose union constitutes the convex hull of the original points. By applying Delaunay triangulation to both dense and sparse point clouds reconstructed using COLMAP, the resulting reconstruction is adaptive; regions near the surface, where point density is higher, are modeled with greater resolution (smaller tetrahedra), while regions farther from the surface are modeled with larger tetrahedra.

Our method leverages adaptive mesh refinement to explicitly handle sparse point clouds. Unlike point cloud-based approaches requiring dense sampling for stable initialization \cite{jager2024hologs}, tetrahedral meshes naturally enforce geometric consistency, even with sparse inputs, via globally coherent topological structures. Compared to triangle meshes \cite{liu2025mesh}, tetrahedral meshes provide greater structural robustness, especially in regions of varying density. The volumetric nature of tetrahedra enables more stable and well-defined interpolation of geometric properties, improving robustness to noise and missing data, and effectively mitigating cumulative errors from local point cloud deficiencies.

After partitioning the domain, we initialize Gaussian components for each face pair of the tetrahedron. To anchor the Gaussian components to the tetrahedral mesh, we align each Gaussian center with the center of a triangular face and parameterize the covariance matrix based on the vertices of the face, denoted as \( V \). For each triangular face \( V \), consisting of three vertices \( v_1 \), \( v_2 \), and \( v_3 \) within the domain \( \mathbb{R}^3 \), we aim to parameterize the Gaussian components by utilizing these vertices. The mean vector is expressed as a convex combination of the vertex positions of \( V \), thereby defining the position \( P \) of each Gaussian component. 
\begin{equation}
    P(\alpha_1,\alpha_2,\alpha_3) = \alpha_1 v_1 + \alpha_3 v_3 + \alpha_3 v_3,
\end{equation}
where $\alpha_1, \alpha_2, \alpha_3$ are trainable parameters, constrained by $\alpha_1 + \alpha_2 + \alpha_3 = 1$. This parameterization ensures that the Gaussian component is always located within the face $V$.

In our experiments, we initialize the number of Gaussian components $k \in \mathbb{N}$ to be placed on each mesh face and define their density over the surface. For a mesh containing $n \in \mathbb{N}$ faces, the final number of Gaussians is fixed and equal to $k \cdot n$.
\subsection{GS Pruning and Structure-aware Densification}
Large-scale scenes contain a wide variety of elements, including various transportation vehicles, buildings, and other structures, coexisting under different lighting conditions. In such cases, 3DGS-based methods require a substantial number of Gaussian functions to model these diverse objects, and the rapid increase in the number of Gaussian components can easily lead to an explosion in disk space requirements. Simply applying 3DGS to large-scale scenes may result in low-quality reconstructions or memory errors due to insufficient GPU memory. Additionally, the large number of Gaussian components during training necessitates sufficient iterations to optimize the entire large-scale scene, which can be time-consuming and potentially unstable.

The 3DGS representation, with its extensive use of Gaussian functions, incurs a significant storage overhead. However, pruning Gaussian functions based on simple criteria (e.g., point opacity) can lead to a substantial decrease in modeling performance, particularly in cases where complex scene structures are removed. Inspired by successful neural network pruning techniques \cite{li2023compressing}, which eliminate less influential neurons without affecting overall network performance, we have customized a general pruning paradigm for point-based representations to reduce the over-parameterized point count while preserving the original accuracy. Therefore, identifying the most representative redundant Gaussians that can be recovered with the desired precision is a crucial step in our approach.

Statistical analysis reveals that most Gaussians have opacity values of either 0 or 1. Directly discarding Gaussian points with low opacity can lead to the removal of a significant number of Gaussians, resulting in substantial distortion in scene representation. Inspired by Equation \ref{prune}, which models the image rendering process by projecting 3D Gaussians onto specific camera viewpoints and rasterizing them, we quantify the importance of each Gaussian based on its contribution to every pixel across all training views, following principles similar to magnitude-based network pruning. During model training, we iterate over all training pixels to compute each Gaussian's hit count, considering both opacity and volume.
\begin{equation} \label{prune}
    GS_j = \sum_{i = 1}^{M H W} I(G(X_j),r_i)\cdot \sigma_j \cdot \gamma_j(\Sigma ),
\end{equation}
where $j$ indicates the Gaussian index, and $M$, $H$, and $W$ represent the number of training views, image height, and width, respectively. $I$ is the indicator function that takes the value of $1$ when a Gaussian intersects with a given ray and $0$ otherwise.

Gaussian distributions are ranked based on global importance scores calculated by the aforementioned formula, and distributions with the lowest scores are proportionally pruned. It is important to note that an excessively high pruning ratio may adversely impact rendering performance, as removing too many Gaussian functions can lead to significant information loss, a severe reduction in scene detail, and a noticeable decline in model accuracy and visual fidelity. While existing point pruning methods, such as LightGaussian \cite{fan2023lightgaussian}, demonstrate potential for storage reduction, our analysis reveals two critical limitations: 1) Their opacity-based importance metric tends to eliminate geometrically critical Gaussians in low-texture regions 2) The purely statistical pruning approach lacks structural awareness, often creating under-sampled areas that degrade rendering quality in complex geometries. To overcome these limitations, we propose a curvature-aware pruning-densification co-design framework.

Inspired by recent research\cite{ververas2024sags}, we observe a prevalent under-representation of low-curvature regions within the scene, leading to artifacts such as blurred rendering. To mitigate this, we introduce a novel structure-aware densification module designed to strategically increase point density in these under-represented, low-curvature areas. This targeted densification results in a more compactly structured model with enhanced expressive capabilities, particularly in regions traditionally prone to rendering deficiencies. The core principle of our approach lies in leveraging the covariance matrix of the local neighborhood to infer surface characteristics.  The core insight is as follows: since the covariance matrix of the local neighborhood can be analyzed to estimate surface characteristics, we leverage the covariance matrix and its eigenvalues to estimate the Gaussian mean curvature of the local region of the point cloud $P$ \cite{Efficient-simplification}.

\begin{equation}
    \rho = \frac{\lambda_0}{\lambda_0 + \lambda_1 + \lambda_2},
\end{equation}
where $\rho$ represents the curvature of the local point cloud $P$. During the training process, we construct a K-nearest neighbor graph for each point. Each point and its neighboring points form a local point cloud $P$, and we calculate the covariance matrix of size $3 \times 3$. $\lambda_i$ (for $i = 0, 1, 2$) represents the eigenvalue of the covariance matrix. The densification process occurs at each low-curvature point in $P$. For this purpose, we construct a mask to mitigate the impact of over-pruning, ensuring fidelity in the low-curvature area.

\subsection{Vector Quantization}
Each Gaussian component contains information including the position \( p \in \mathbb{R}^3 \), spherical harmonic coefficients \( sh \in \mathbb{R}^{(k + 1)^2 \times 3} \) (where \( k \) denotes the order, resulting in 48 parameters for \( k=3 \)), opacity \( \alpha \in \mathbb{R} \), rotation quaternion \( r \in \mathbb{R}^4 \), and scaling factor \( s \in \mathbb{R}^3 \). Notably, 3DGS representations typically require millions of Gaussians to model a scene, leading to gigascale training models that impose significant storage requirements, a challenge that is particularly pronounced in unbounded outdoor scenes. We observe that numerous Gaussian may exhibit similar parameter values (e.g., covariance), thus facilitating the compression of the final model based on the principle of vector quantization.

\section{Experiments}
\subsection{Datasets}
We evaluate our method against state-of-the-art methods on 13 outdoor scenes of varying scales, including eight from four publicly available datasets and five self-collected scenes.

{\bf{Public Scenes.}} The Mill 19 Dataset \cite{turki2022mega}, which consists of two scenes: Rubble and Building, captured near a former industrial site. The Building scene encompasses a rectangular area of 500 × 250 square meters around a large industrial structure, while the Rubble scene covers a nearby construction zone. Each scene contains over 1,000 multi-view images at 4K resolution. The MatrixCity Dataset \cite{li2023matrixcity} is a large-scale, high-quality synthetic urban dataset. We selected a small urban scene covering an area of 2.7 square kilometers for our experiments, aiming to evaluate the model's performance on synthetic datasets. The WHU Dataset \cite{whu} is a synthetic aerial dataset designed for large-scale reconstruction of the Earth's surface, covering an area of 6.7 × 2.2 square kilometers. This area includes dense high-rises, sparse factories, forested mountains, as well as exposed ground and rivers. The dataset is divided into six distinct sub-scenes, and in our experiments, we selected areas 1, 4, and 5 for evaluation. The Tanks \& Temples Dataset \cite{knapitsch2017tanks} has the train and truck scenes to evaluate our experiments, assessing the performance of our model in small outdoor scenarios.

{\bf{Self-collected Scenes.}} To further validate our method's performance in campus environments, we collect the SCUT Campus Aerial (SCUT-CA) scenes, which include three scenes featuring teaching, office and history museum buildings. Specifically, we capture images from approximately 100 meters above each of the three distinct scenes using the DJI Matrice 300 RTK, resulting in a total of 125, 197, and 273 images, respectively, for an overall total of 595 images. Each image has a resolution of 4082 × 2148 pixels. The dataset encompasses a variety of architectural styles and scene types, which can intuitively demonstrate the performance of our method in real-world reconstruction.To ensure consistency with the 3DGS input, we downsample the images to a resolution of 1600x841. We also collect the plateau region scenes, which comprises two scenes characterized by weak and repetitive textures, along with diverse geographical features. Specifically, we captured images in the plateau region of Tibet at an altitude of approximately 100 meters, acquiring 203 and 170 images, respectively, for a total of 373 images. Each image has a resolution of 6222 × 4148 pixels, and we also downsampled these images to a resolution of 1600 × 841 pixels. Camera parameters were estimated using widely adopted techniques such as COLMAP \cite{sfm}.
\subsection{Evaluation Metrics}
To comprehensively evaluate the reconstruction quality of different methods, we adopt standard metrics, including SSIM, PSNR, and LPIPS, as our evaluation indicators.
\subsection{Implementation Details}
All experiments are conducted on an NVIDIA RTX 3090 24GB GPU using the PyTorch framework. We downsample the images and use a unified resolution of 1.6k for training. To ensure a fair comparison of the experimental results, our model typically requires 1 to 2 hours to optimize for large scenes, depending on factors such as scene complexity, dataset characteristics, the number of Gaussians per face, and the number of vertices. For the initial number of Gaussians per facet, we generally adopt the default value of \( k = 3 \). We set the total number of training iterations to 30,000 and apply pruning and structure-aware densification strategies at 12,000 and 20,000 iterations. During training, we quantize the color, spherical harmonics, scale, and rotation parameters separately using K-Means algorithm. To perform compression comparisons, we trained two models, one of which utilized a codebook for compression. In the comparative experiments, we analyze various metrics of the models both before and after compression.
\subsection{Results}
\subsubsection{Results on Mill 19 and MatrixCity Dataset}
\begin{figure*}
    \centering
    \includegraphics[width=4.0 in]{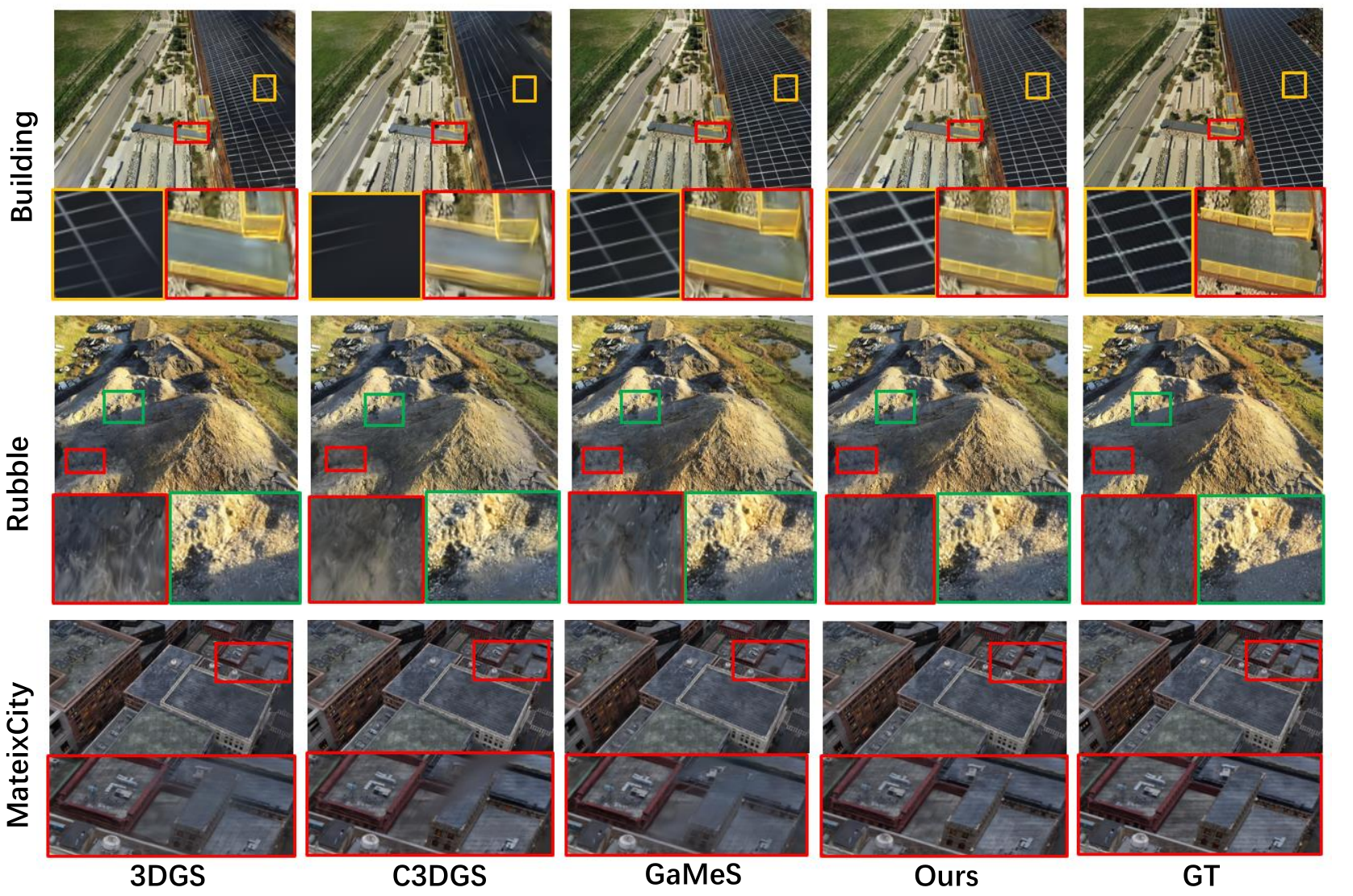}
    \caption{Qualitative comparison of our method with other competitive methods on the Mill 19 and MatrixCity datasets. \textcolor{red}{Red}, \textcolor{yellow}{yellow}, and \textcolor{green}{green} squares highlight the visual differences for clearer comparison. Our model achieves more refined reconstruction, as evidenced in the Building scene. Our model excels in rendering quality, particularly in low-semantic regions such as grid lines.}
    \label{fig:mill-19}
\end{figure*}
\begin{table*}[!t] 
\caption{Quantitative Comparison on Mill 19 and MatrixCity large-scale scene datasets. The '-' symbol indicates Mega-NeRF \cite{turki2022mega} and Switch-NeRF \cite{zhenxing2022switch} were not evaluated on \textit{MatrixCity} due to difficulties in adjusting its training configurations beyond the provided, resulting in poor performance on this dataset. The best results of each metric are in \textbf{bold}.}
  \label{tab:table1}
  \centering
  \tabcolsep=0.1cm
  \resizebox{0.72\textwidth}{!}{%
  \begin{tabular}{@{}l|ccc|ccc|ccc@{}}
    \hline
     & \multicolumn{3}{c|}{MatrixCity} & \multicolumn{3}{c|}{Rubble} & \multicolumn{3}{c}{Building} \\
    \hline
    Metrics & SSIM$\uparrow$ & PSNR$\uparrow$ & LPIPS$\downarrow$ & SSIM$\uparrow$ & PSNR$\uparrow$ & LPIPS$\downarrow$ & SSIM$\uparrow$ & PSNR$\uparrow$ & LPIPS$\downarrow$ \\
    \hline
    MegaNeRF \cite{turki2022mega} & - & - & - & 0.553 & 24.06 & 0.516 & 0.547 & 20.93 & 0.504 \\
    Switch-NeRF \cite{zhenxing2022switch} & - & - & - & 0.562 & 24.31 & 0.496 & 0.579 & 21.54 & 0.425 \\
    C3DGS \cite{compact-3dgs} & 0.795 & 26.03 & 0.370 & 0.719 & 23.00 & 0.372 & 0.645 & 20.00 & 0.425\\
    3DGS \cite{3dgs} & 0.797 & 25.16 & 0.343 & 0.746 & 24.30 & 0.324 & 0.676 & 20.11 & 0.372 \\
    GaMes \cite{waczynska2024games} & 0.773 & 24.64 & 0.367 & 0.767 & 24.64 & 0.311 & 0.698 & 20.97 & 0.329 \\
    Scaffold-GS \cite{lu2024scaffold} & \textbf{0.842} & \textbf{27.25} & \textbf{0.290} & 0.736 & 23.71 & 0.346 & 0.711 & 21.55 & 0.328 \\
    \hline
    Our(Quantify) & 0.788 & 24.96 & 0.342 & 0.782 & 24.82 & 0.292 & 0.712 & 21.32 & 0.318 \\
    Ours & 0.792 & 25.28 & 0.337 & \textbf{0.788} & \textbf{25.19} & \textbf{0.287} & \textbf{0.720} & \textbf{21.58} & \textbf{0.314} \\
    \hline
  \end{tabular}
  }
\end{table*}
In Table. \ref{tab:table1}, we report the SSIM, PSNR, and LPIPS metrics for the Mill 19 and MatrixCity scenes. Our EA-3DGS method outperforms comparative methods in all aspects for the outdoor large-scale scene in Mill 19, showing greater advantages in SSIM and LPIPS metrics. This indicates that our method is better at perceiving the structural information of large-scale scenes, rendering, and reconstructing richer scene details. In Fig. \ref{fig:mill-19}, we provide a comparison of scene effects with other methods based on 3DGS. For the Mill 19 dataset, our method achieves visually pleasing renderings, with improved detail in features such as eaves and tiles compared to other methods. However, under the MatrixCity urban street scene dataset, our method's results are inferior to those of the Scaffold-GS method. This may be due to the complexity of urban housing structures posing challenges for generating efficient meshes, which affects the initialization of Gaussian components and leads to poor reconstruction results in regions with occlusions and low semantic content.
\subsubsection{Results on SCUT-CA Scenes}
\begin{table*}[!t]
\caption{Quantitative Comparison on SCUT-CA Scenes. The best results of each metric are in \textbf{bold}.}
  \label{tab:table2}
  \centering
  \tabcolsep=0.1cm
  \resizebox{0.72\textwidth}{!}{%
  \begin{tabular}{@{}l|ccc|ccc|ccc@{}}
    \hline
     & \multicolumn{3}{c|}{Tower Two} & \multicolumn{3}{c|}{School History Museum} & \multicolumn{3}{c}{EI School} \\
    \hline
    Metrics & SSIM$\uparrow$ & PSNR$\uparrow$ & LPIPS$\downarrow$ & SSIM$\uparrow$ & PSNR$\uparrow$ & LPIPS$\downarrow$ & SSIM$\uparrow$ & PSNR$\uparrow$ & LPIPS$\downarrow$ \\
    \hline
    3DGS \cite{3dgs} & 0.887 & 29.87 & 0.143 & 0.849 & 28.54 & 0.189 & 0.882 & 29.12 & 0.132 \\
    GaMeS \cite{waczynska2024games} & 0.891 & 29.99 & 0.140 & 0.861 & 28.83 & 0.182 & 0.889 & 29.36 & 0.120 \\
    C3DGS \cite{compact-3dgs} & 0.873 & 29.03 & 0.164 & 0.844 & 27.96 & 0.211 & 0.863 & 27.97 & 0.160 \\
    Scaffold-GS \cite{lu2024scaffold} & 0.880 & 29.49 & 0.155 & 0.854 & 28.44 & 0.197 & 0.873 & 28.67 & 0.147 \\
    \hline
    Our(Quantify) & 0.895 & 30.27 & 0.134 & 0.864 & 28.96 & 0.176 & 0.892 & 29.43 & 0.119 \\
    Ours & \textbf{0.902} & \textbf{30.53} & \textbf{0.130} & \textbf{0.881} & \textbf{29.33} & \textbf{0.172} & \textbf{0.904} & \textbf{29.71} & \textbf{0.111} \\
    \hline
  \end{tabular}
  }
\end{table*}

\begin{figure*}
    \centering
    \includegraphics[width=4.0 in]{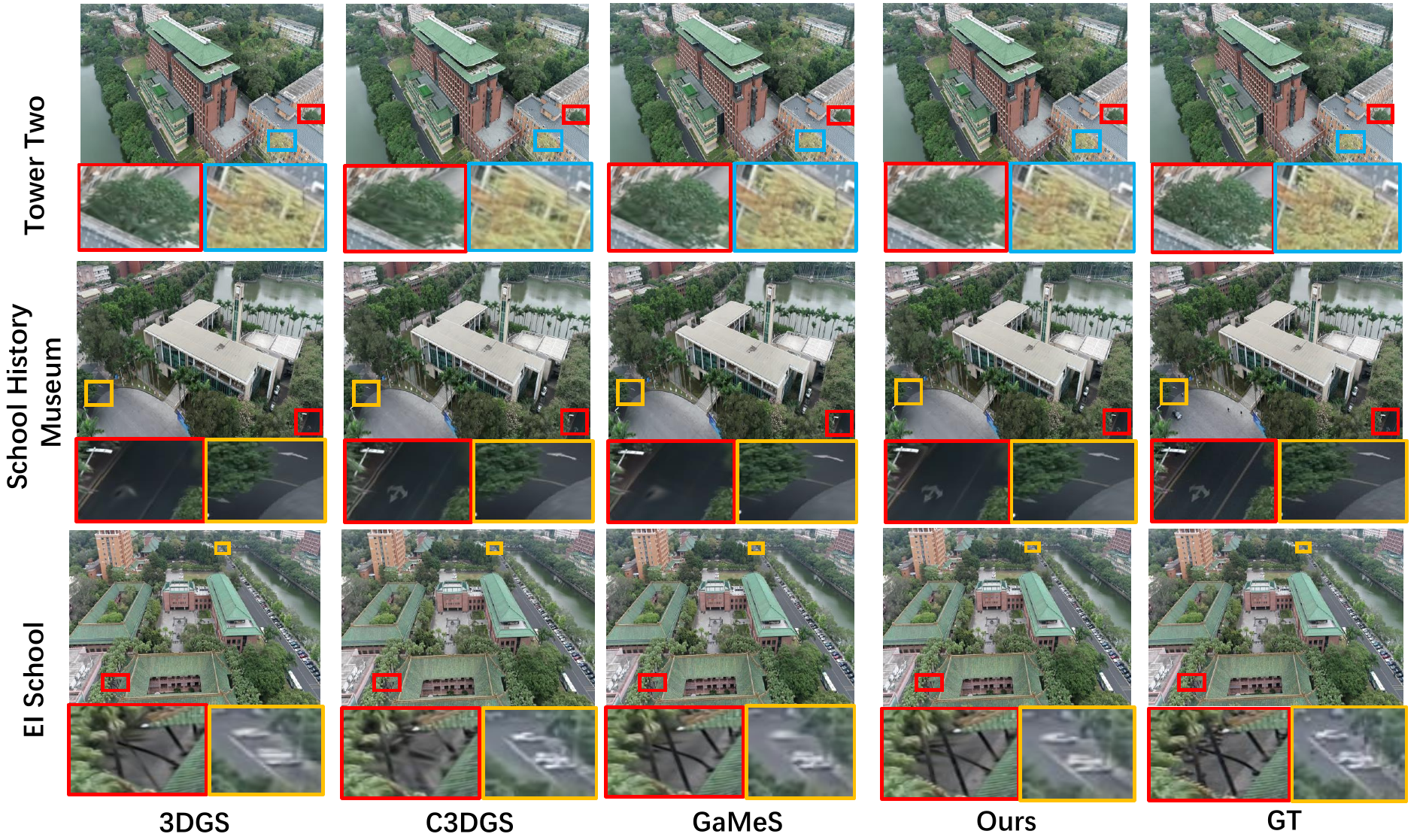}
    \caption{Qualitative comparison of our method with other competitive methods on the SCUT-CA scenes. \textcolor{red}{Red}, \textcolor{yellow}{yellow}, and \textcolor{cyan}{blue} squares highlight the visual differences for clearer comparison. Our method demonstrates superior fidelity in handling details, which is particularly evident in the School History Museum scene. Specifically, our method achieves more accurate reconstruction of road lane markings and directional signs.}
    \label{fig:scut}
\end{figure*}

In Table. \ref{tab:table2}, we present the SSIM, PSNR, and LPIPS metrics for the SCUT-CA scenes. Our EA-3DGS method outperforms comparative methods in all aspects of this dataset, which includes numerous high-rise buildings and trees distributed throughout the campus scenes. Our model demonstrates more structured and refined characteristics, enabling the effective presentation of scene details. In the tower-two scene, our method achieves a PSNR of 30.53 and surpasses other methods in both SSIM and LPIPS. A visual comparison in Fig. \ref{fig:scut} illustrates that our images exhibit fewer artifacts during rendering and provide better restoration of details such as marking lines, indicating that our method demonstrates superior performance in outdoor multi-story building scenes.
\subsubsection{Results on WHU dataset and Plateau Scenes}
\begin{table*}[!t]
\caption{Quantitative Comparison on WHU dataset and Plateau Scenes. The best results of each metric are in \textbf{bold}.}
  \label{tab:table3}
  \centering
  \tabcolsep=0.1cm
  \resizebox{0.90\textwidth}{!}{%
  \begin{tabular}{@{}l|ccc|ccc|ccc|ccc|ccc@{}}
    \hline
     & \multicolumn{9}{c|}{WHU} & \multicolumn{6}{c}{Plateau} \\
    \hline
     & \multicolumn{3}{c|}{Area1} & \multicolumn{3}{c|}{Area4} & \multicolumn{3}{c|}{Area5} & \multicolumn{3}{c|}{Scene1}& \multicolumn{3}{c}{Scene2} \\
    \hline
    Metrics & SSIM$\uparrow$ & PSNR$\uparrow$ & LPIPS$\downarrow$ & SSIM$\uparrow$ & PSNR$\uparrow$ & LPIPS$\downarrow$ & SSIM$\uparrow$ & PSNR$\uparrow$ & LPIPS$\downarrow$& SSIM$\uparrow$ & PSNR$\uparrow$ & LPIPS$\downarrow$& SSIM$\uparrow$ & PSNR$\uparrow$ & LPIPS$\downarrow$ \\
    \hline
    3DGS \cite{3dgs} & 0.887 & 30.84 & 0.216 & 0.901 & 28.87 & 0.200 & 0.799 & 26.04 & 0.329  & 0.723 & 28.46  & 0.334  & 0.745 & 29.62  & 0.369 \\
    GaMes \cite{waczynska2024games} & 0.902 & 31.28 & 0.207 & 0.892 & 28.34 & 0.203 & 0.812 & 26.48 & 0.311 & 0.783 & 29.24 & 0.308  & 0.759 & 29.79  & 0.363  \\
    C3DGS \cite{compact-3dgs} & 0.922 & 32.69 & 0.217 & 0.946 & 33.79 & 0.147 & 0.865 & 28.37 & 0.273 & 0.721 & 28.32 & 0.355 & 0.721 & 29.78 & 0.400 \\
    Scaffold-GS \cite{lu2024scaffold} & \textbf{0.978} & \textbf{38.90} & \textbf{0.068} & \textbf{0.980} & \textbf{38.86} & \textbf{0.063} & \textbf{0.964} & \textbf{34.70} & \textbf{0.089} & 0.787 & 29.66 & 0.308 & 0.737 & 30.06 & 0.393 \\
    \hline
    Our(Quantify) & 0.913 & 31.56 & 0.204 & 0.907 & 29.65 & 0.196 & 0.826 & 26.61 & 0.305 & 0.814 & 30.67 & 0.290 &  0.818 & 31.07 & 0.331 \\
    Ours & 0.920 & 31.81 & 0.199 & 0.919 & 30.34 & 0.189 & 0.832 & 26.97 & 0.298 &  \textbf{0.827} & \textbf{31.19} & \textbf{0.278} & \textbf{0.833} & \textbf{31.68} & \textbf{0.324} \\
    \hline
  \end{tabular}
  }
\end{table*}
\begin{figure*}
    \centering
    \includegraphics[width=4.0 in]{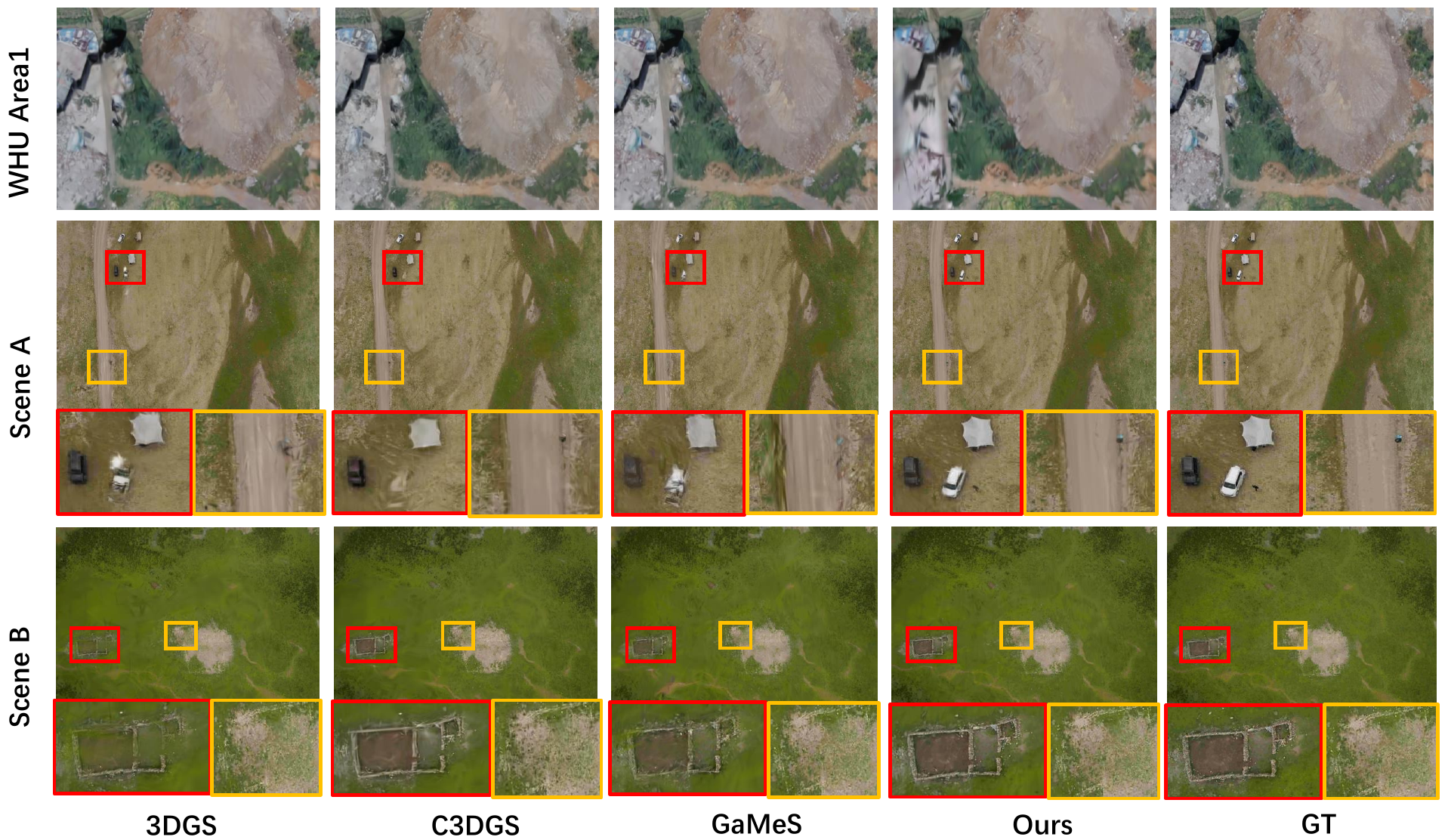}
    \caption{Qualitative comparison of our method with other competitive methods on the WHU and Plateau large-scale scene datasets. \textcolor{red}{Red} and \textcolor{yellow}{yellow} squares highlight the visual differences for clearer comparison. Our method demonstrates superior reconstruction quality. In scene A of the Plateau dataset, our method effectively reconstructs the details of the car and tent captured from high altitude, aligning more closely with the real scene.}
    \label{fig:whu}
\end{figure*}

In Table. \ref{tab:table3}, we present the SSIM, PSNR, and LPIPS metrics for the WHU and Plateau scenes. Our EA-3DGS method demonstrates superior performance compared to comparative methods across all metrics in the Plateau dataset, which was captured in a plateau environment and poses significant challenges for texture modeling. Our model excels in representing open areas and effectively reconstructing geometric structures through mesh constraints, resulting in images with enhanced detail. As illustrated in Fig. \ref{fig:whu}, the comparison results indicate that our model outperforms others in both texture fidelity and overall light perception. Conversely, our model underperforms in the WHU scene, likely due to the low resolution of the images in the WHU dataset, which impedes efficient mesh extraction and leads to a considerable number of artifacts in the rendering outputs.
\subsubsection{Results on Tanks and Temples Dataset}
\begin{table}[!t]
\caption{Quantitative Comparison on Tanks and Temples datasets. The best results of each metric are in \textbf{bold}.}
  \label{tab:table4}
  \centering
  \tabcolsep=0.1cm
  \resizebox{0.60\textwidth}{!}{%
  \begin{tabular}{@{}l|ccc|ccc@{}}
    \hline
     & \multicolumn{3}{c|}{Train} & \multicolumn{3}{c}{Truck} \\
    \hline
    Metrics & SSIM$\uparrow$ & PSNR$\uparrow$ & LPIPS$\downarrow$ & SSIM$\uparrow$ & PSNR$\uparrow$ & LPIPS$\downarrow$ \\
    \hline
    3DGS \cite{3dgs} & 0.816 & 22.09 & 0.206 & 0.873 & 25.11 & 0.154  \\
    GaMeS \cite{waczynska2024games} & 0.816 & 22.11 & 0.205 & 0.877 & 25.24 & 0.151\\
    C3DGS \cite{compact-3dgs} & 0.792 & 21.56 & 0.240 & 0.871 & 25.07 & 0.163 \\
    Scaffold-GS \cite{lu2024scaffold} & 0.816 & 22.32 & 0.213 & 0.886 & 25.83 & 0.141 \\
    \hline
    Our(Quantify) & 0.819 & 22.34 & 0.201 & 0.879 & 25.46 & 0.146 \\
    Ours & \textbf{0.828} & \textbf{22.91} & \textbf{0.191} & \textbf{0.887} & \textbf{25.93} & \textbf{0.138} \\
    \hline
  \end{tabular}
  }
\end{table}

In Table. \ref{tab:table4}, we provide the SSIM, PSNR, and LPIPS metrics for the Tanks and Temples datasets. This dataset focuses on capturing images of trains and trucks, with the modeling challenge primarily centered around background reconstruction. Our EA-3DGS method outperforms comparative methods across all metrics. 

\subsection{Ablation Study}
In this section, we use the Mill 19 dataset to evaluate the effectiveness of each component of our method and the rationality of the model parameter settings.
\subsubsection{Effect of Tetrahedra Mesh Initialization}
\begin{table}[!h]
\centering
\caption{Ablation study on Curvature perception densification and mesh selection \label{tab:ab_1}}
\begin{tabular}{l|cccc}
\toprule
method & SSIM $\uparrow$ & PSNR $\uparrow$ &LPIPS $\downarrow$ & Size $\downarrow$ \\
\midrule
\hspace{5.0mm}-& 0.711 & 22.20 & 0.348  & 1.9GB \\ 
w tri-mesh & 0.723 & 22.58 & 0.340  & 1.7GB \\ 
w tetra-mesh & 0.750 & 23.27 & 0.304  & 1.8GB \\
w curv.-aware. & \textbf{0.754} & \textbf{23.39} & \textbf{0.301}  & 1.8GB \\
w codebook & 0.747 & 23.07 & 0.305 & \textbf{0.3GB} \\ 
\bottomrule
\end{tabular}
\end{table}
\begin{figure}
    \centering
    \includegraphics[width=0.7\linewidth]{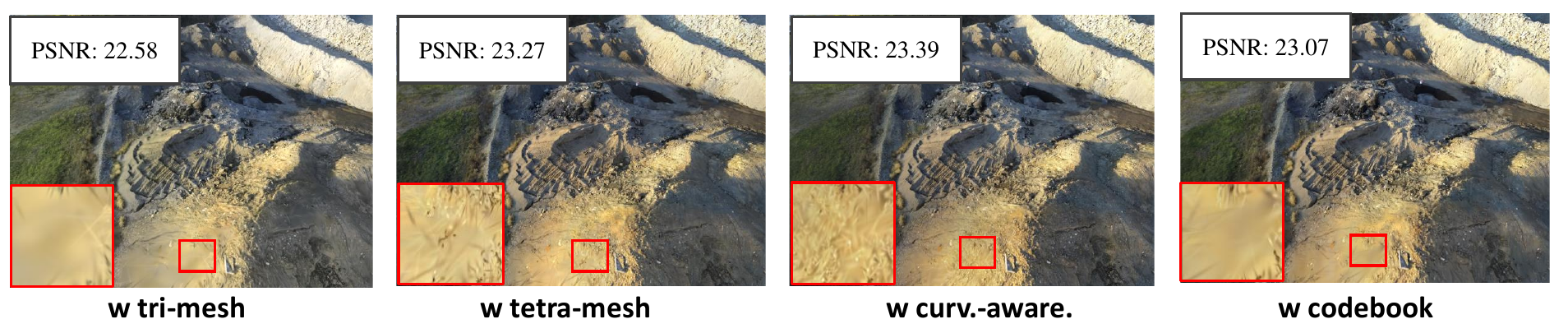}
    \caption{Ablation study on the components of EA-3DGS. We conduct a series of ablation experiments on the Mill 19 datasets, presenting qualitative results from the rubble scenes. We emphasize the differences across model configurations using the same crop of the resulting images and highlight visual differences with \textcolor{red}{red} squares.
}
    \label{fig:ab1}
\end{figure}
\begin{table}[!h]
\centering
\caption{Ablation study on  number of splatting per face $k$ \label{tab:ab_2}}
\begin{tabular}{l|ccc}
\toprule
method & SSIM $\uparrow$ & PSNR $\uparrow$ & LPIPS $\downarrow$ \\
\midrule
\hspace{2.5mm}- & 0.711 & 22.20 & 0.348 \\
k = 1 & 0.717 & 22.68 & 0.341 \\
k = 2 & 0.732 & 23.10 & 0.321 \\
k = 3 & \textbf{0.754} & \textbf{23.39} & \textbf{0.301} \\
k = 4  & 0.746 & 23.28 & 0.315 \\
\bottomrule
\end{tabular}
\end{table}
\begin{figure}
    \centering
    \includegraphics[width=0.7\linewidth]{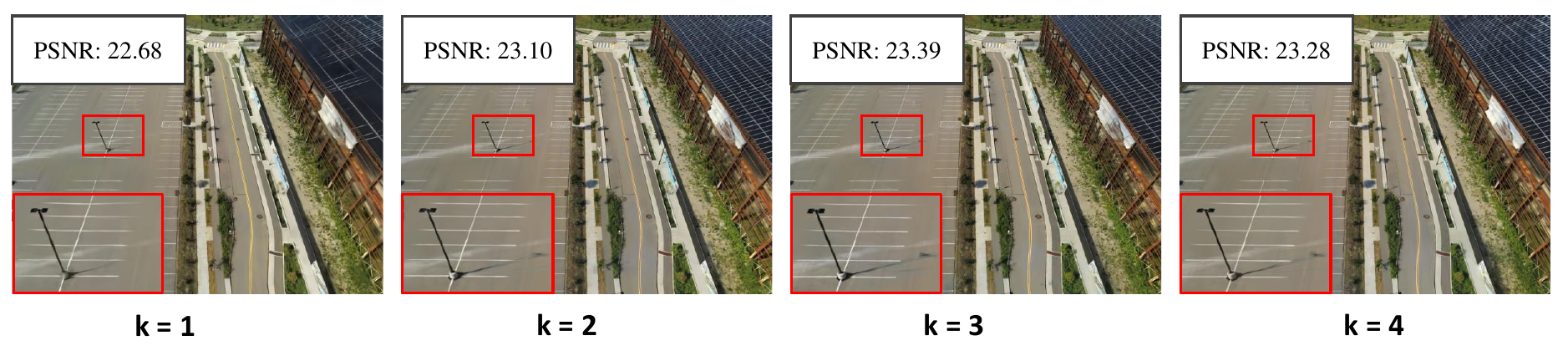}
    \caption{Ablation study on the parameter settings of EA-3DGS. We analyze the impact of hyperparameter \( k \) (initial number of Gaussians per mesh face) in EA-3DGS through experiments on the Mill 19 dataset. Focusing on building scenes, we compare qualitative results across varying \( k \). Visual distinctions are emphasized by consistent image crops, with key differences highlighted in \textcolor{red}{red} boxes.}
    \label{fig:ab2}
\end{figure}

In the comparative analysis presented in Fig. \ref{fig:ab1} and Table. \ref{tab:ab_1}, we demonstrate the significant visual differences of our proposed model when employing conventional SFM techniques versus tetrahedral mesh initialization methods. The experimental results clearly indicate that the tetrahedral mesh initialization strategy exhibits superior reconstruction capabilities when the Gaussian model is applied to low-texture regions. This approach not only enhances the reconstruction accuracy in low-texture areas but also effectively reduces artifacts across the entire scene, thereby improving the overall visual quality. In contrast, conventional methods, which rely on precise SFM initialization, lack certain structural characteristics.
\subsubsection{Effect of Curvature perception densification}
In the detailed comparative analysis presented in Fig. \ref{fig:ab1} and Table. \ref{tab:ab_1}, we examine the visual differences of our proposed method with and without curvature densification.  In large-scale outdoor environments, which are highly challenging, an effective densification strategy can significantly enhance our model's ability to fill low-texture regions, especially immediately following the initial cropping phase.  The results demonstrate that curvature-based densification can effectively improve the overall scene rendering quality. 
\subsubsection {Effect of model parameter setting}
In the comparative analysis presented in Fig. \ref{fig:ab2} and Table \ref{tab:ab_2}, we included experimental results for the hyperparameter $k$, which represents the initial number of Gaussians on each face of the mesh. Undoubtedly, the rendering performance of the 3DGS method is significantly influenced by the initialization of Gaussians. In our approach, we leverage a well-structured mesh to initialize the positions of Gaussians. For the setting of the number of Gaussians $k$ on each face, we ultimately selected $3$ based on considerations of convergence speed and rendering quality.
\section{Conclusion and future work}
In this paper, we present a high-quality method designed for real-time rendering of outdoor scenes. Our method introduces an adaptive representation of tetrahedral meshes to partition the scene and initializes 3D Gaussians on each mesh face, thereby enabling robust rendering even in low-texture regions. During the training process, we propose an efficient Gaussian pruning strategy to reduce memory usage by evaluating each 3D Gaussian's contribution to the view and pruning it accordingly. To mitigate potential negative impacts of pruning while preserving structurally significant points, we introduce a structure-aware densification strategy that selectively increases Gaussian density in low-curvature regions. To further enable real-time rendering in larger scenes, we leverage the codebook technique to implement parameter quantization for Gaussian components, effectively reducing disk space overhead. Extensive experiments demonstrate that the proposed method outperforms existing mainstream methods, achieving high-fidelity and fast rendering across multiple outdoor scene datasets.

\textbf{Limitations}: While our EA-3DGS effectively captures the structural features of scenes, it faces challenges in generating effective meshes for datasets with blurry views, resulting in suboptimal performance. As evidenced by the results on the WHU dataset, this motivates us to further optimize our method.

\bibliographystyle{elsarticle-num}
\bibliography{reference}
\end{document}